\documentclass[letterpaper, 10 pt, conference]{ieeeconf}

\IEEEoverridecommandlockouts
\usepackage[english]{babel}
\usepackage{amsmath}  % assumes amsmath package installed
\usepackage{amssymb}  % assumes amsmath package installed
\usepackage{graphicx}
\usepackage{subfigure}
\usepackage{caption}
\usepackage{multirow}
\usepackage{amsfonts}
\usepackage{tabularx}
\usepackage{algorithm,algorithmic}
\usepackage{bm}
\usepackage{breakurl}
\usepackage{enumerate}
\usepackage{adjustbox}
\usepackage{xcolor}
\usepackage{siunitx}
\newcolumntype{b}{X}
\newcolumntype{s}{>{\hsize=.3\hsize}X}
\newcolumntype{x}{>{\hsize=.5\hsize}X}
\usepackage{tikz}
\usetikzlibrary{shapes.geometric, arrows}
\usepackage{array}
\usepackage{cite}
\usepackage{wasysym}
\usepackage{url}
\usepackage{xurl}
\usepackage{booktabs}
\makeatletter
\let\NAT@parse\undefined
\makeatother
\usepackage{hyperref}
\usepackage{cleveref}

\tikzstyle{startstop} = [rectangle, rounded corners, minimum width=3cm, minimum height=1cm,text centered, draw=black, fill=red!30]
\tikzstyle{arrow} = [thick,->,>=stealth]

\title{\LARGE \bf AutoCharge: Autonomous Charging for Perpetual Quadrotor Missions}

\author{Alessandro Saviolo$^{1*}$, Jeffrey Mao$^{1*}$, Roshan Balu T M B$^{1*}$, Vivek Radhakrishnan$^{12}$, and Giuseppe Loianno$^{1}$
\thanks{$^*$These authors contributed equally.}
\thanks{$^1$The authors are with the New York University, Tandon School of Engineering, Brooklyn, NY 11201, USA. {\tt\footnotesize email: \{as16054, jm7752, rt2420, loiannog\}@nyu.edu}.}
\thanks{$^2$The author is with the Autonomous Robotics Research Center-Technology Innovation Institute, Abu Dhabi, UAE. {\tt\footnotesize email:  vivek.radhakrishnan@tii.ae}.}
\thanks{This work was supported by the NSF CAREER Award 2145277, the DARPA YFA Grant D22AP00156-00, Qualcomm Research, Nokia, and NYU Wireless. Giuseppe Loianno
serves as consultant for the Technology Innovation Institute. This arrangement
has been reviewed and approved by the New York University in accordance
with its policy on objectivity in research.}
}

\begin{document}

\makeatletter
\g@addto@macro\@maketitle{
\setcounter{figure}{0}
    \centering
    \includegraphics[width=0.19\linewidth, angle=-90, trim=75 40 250 75, clip] {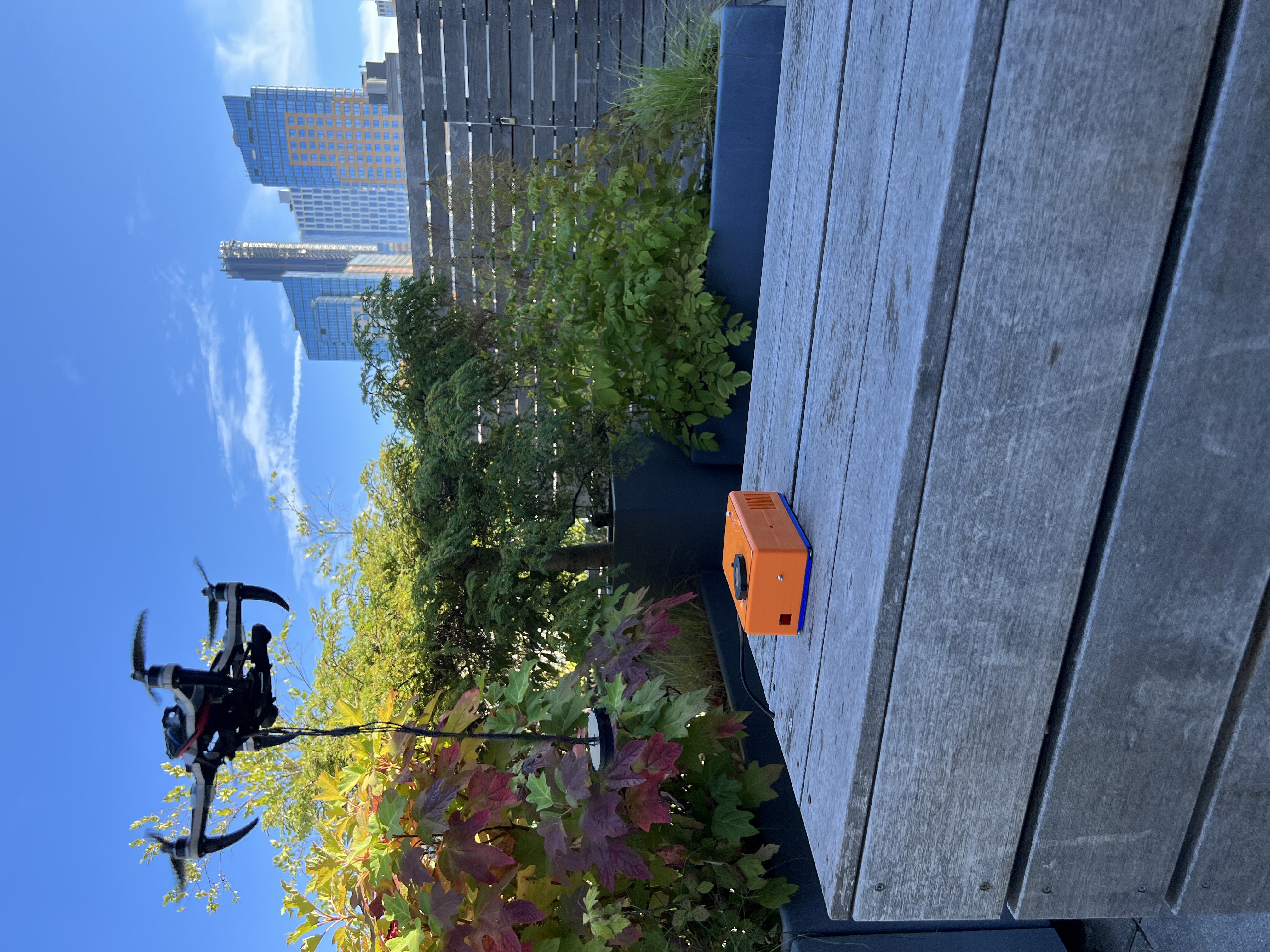}
    \includegraphics[width=0.19\linewidth, angle=-90, trim=75 40 250 75, clip] {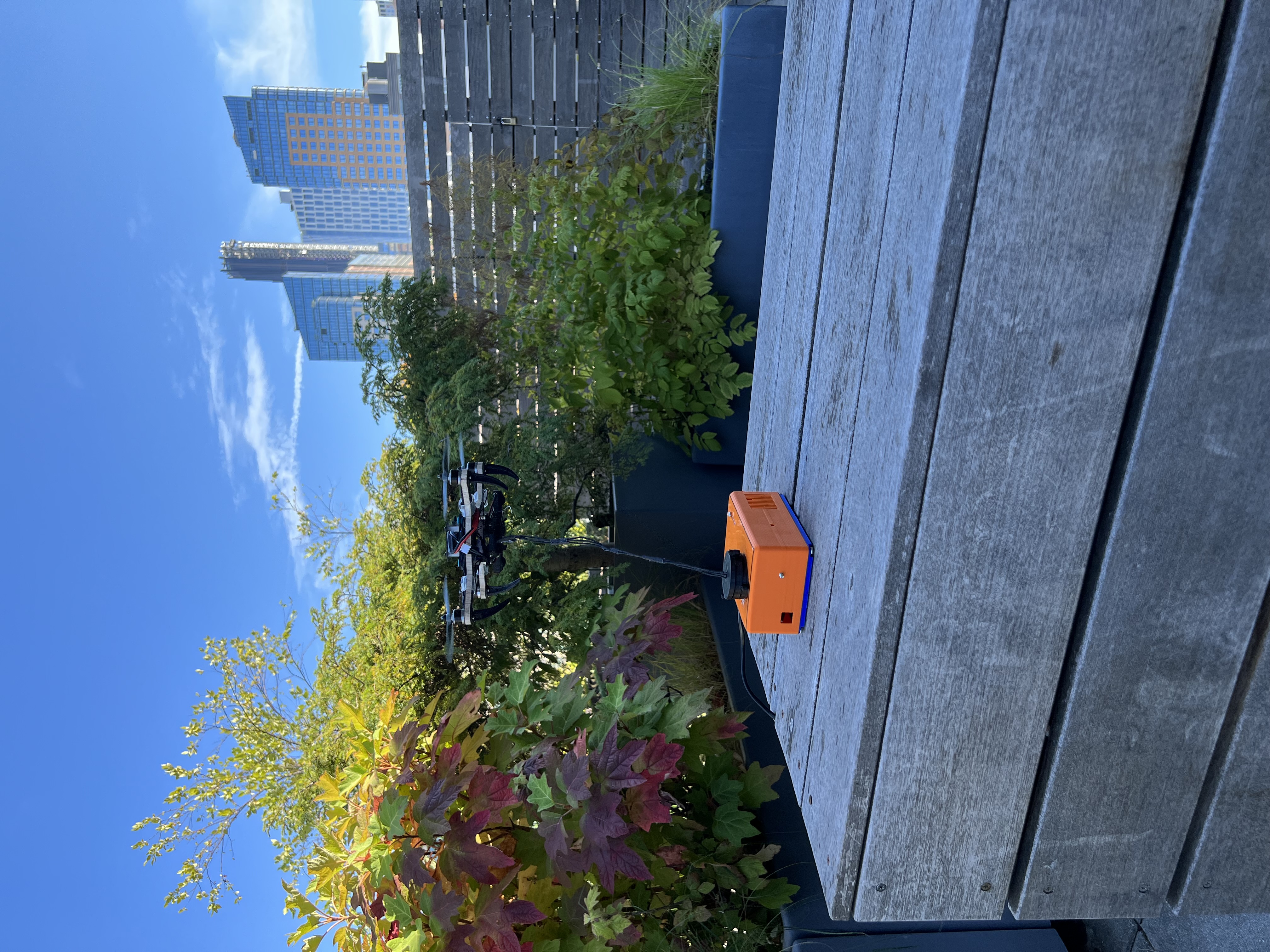}
    \includegraphics[width=0.19\linewidth, angle=-90, trim=75 40 250 75, clip] {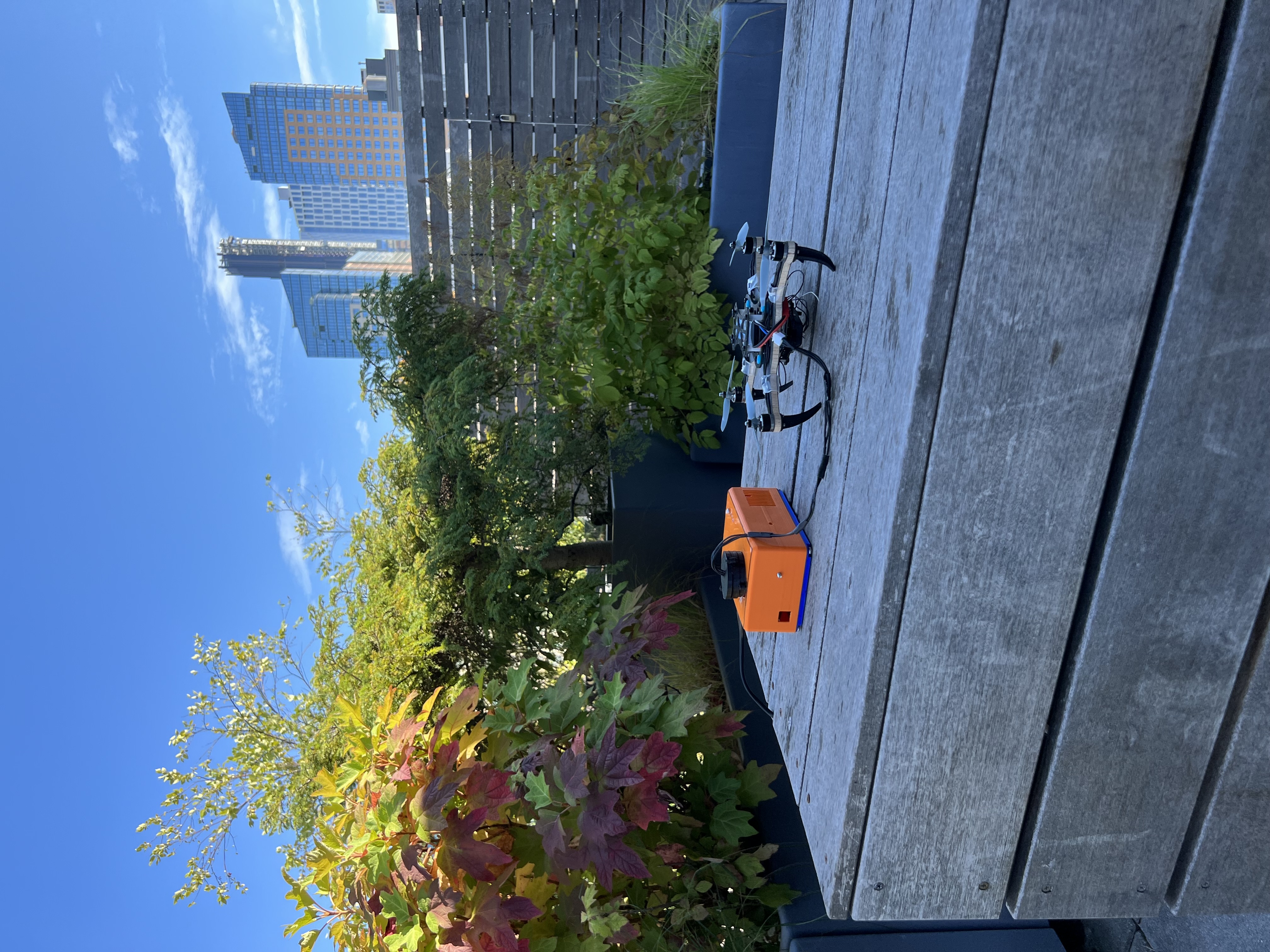}
    \includegraphics[width=0.19\linewidth, angle=-90, trim=75 40 250 75, clip] {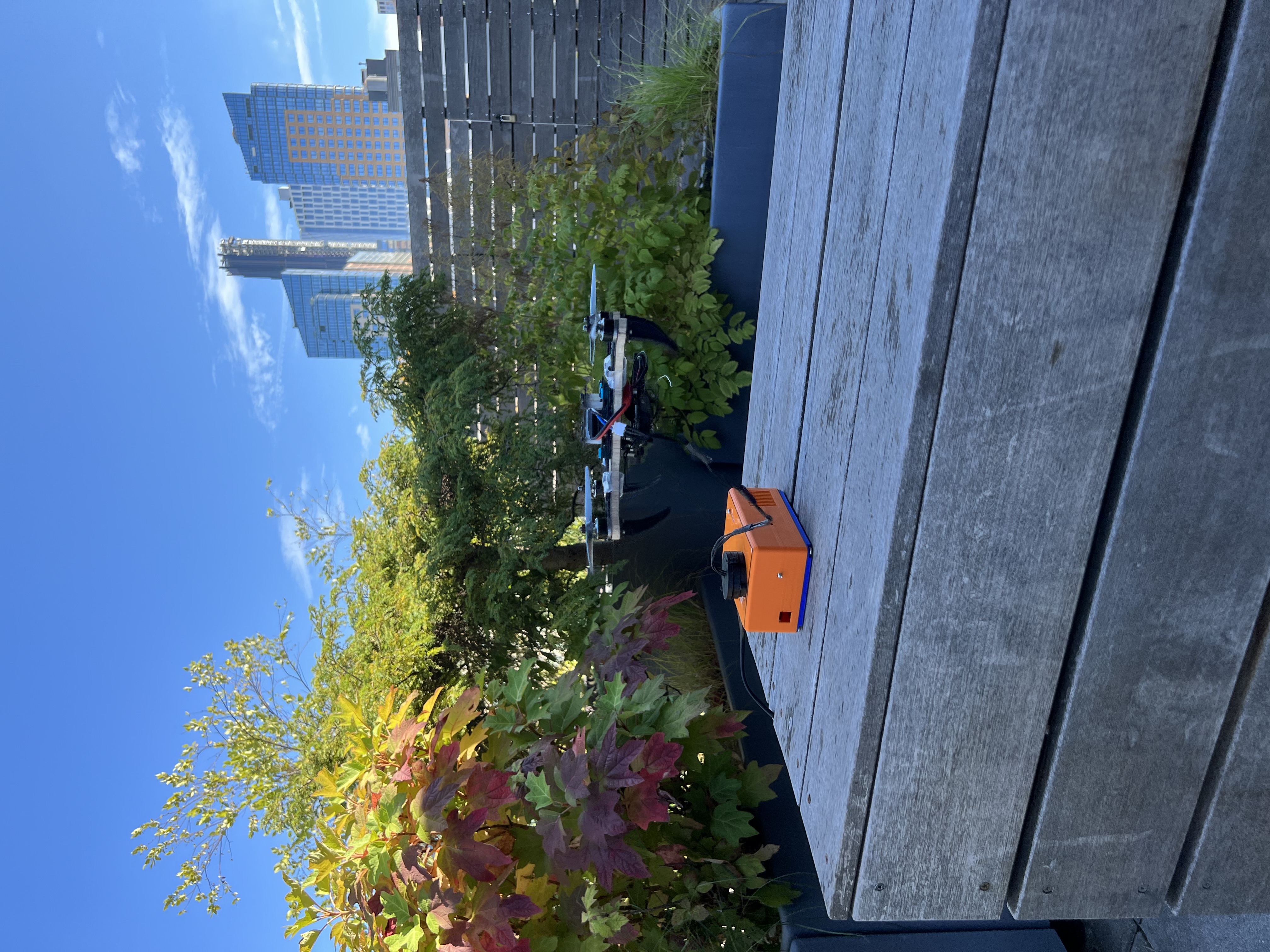}
    \includegraphics[width=0.19\linewidth, angle=-90, trim=75 40 250 75, clip] {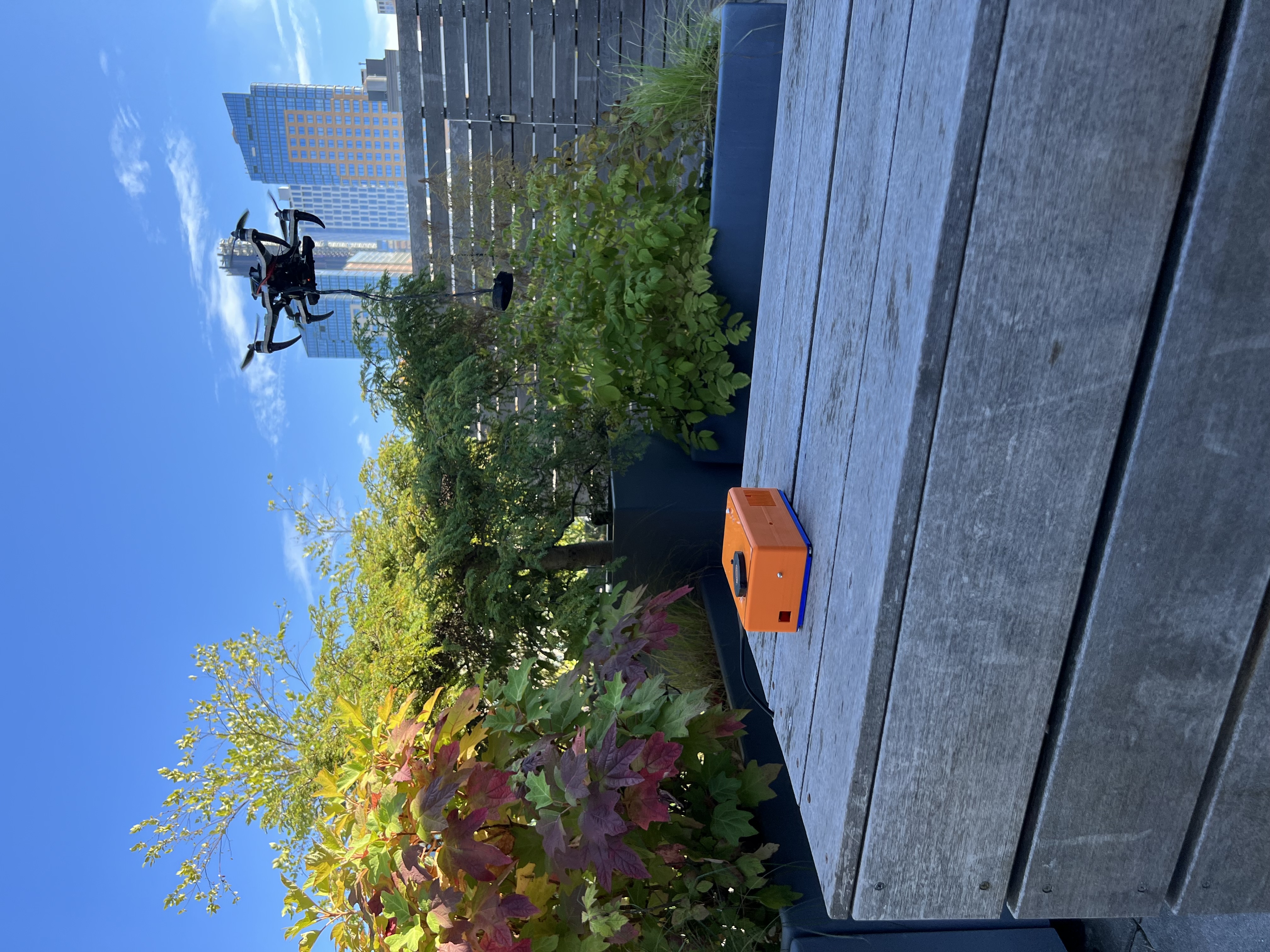}
    \captionof{figure}{
    AutoCharge's operating principle. By leveraging circular magnetic connectors and electromagnets, the proposed charging system ensures solidly repeatable docking (a-b) and un-docking (d-e), enabling perpetual flight missions.
    % Autonomous charging of quadrotor on outdoor flight. (a-b) Magnetic connectors and an electro-permanent magnet ensure precise electrical connection while docking; (c) Tethered charging enables high efficiency charging with minimum power dissipation; (d-e).
    \label{fig:charging_pipeline}}
    \vspace{-0.5em}
    }
\makeatother

\maketitle
\thispagestyle{empty}
\pagestyle{empty}

\begin{abstract}
Battery endurance represents a key challenge for long-term autonomy and long-range operations, especially in the case of aerial robots.
In this paper, we propose AutoCharge, an autonomous charging solution for quadrotors that combines a portable ground station with a flexible, lightweight charging tether and is capable of universal, highly efficient, and robust charging.
We design and manufacture a pair of circular magnetic connectors to ensure a precise orientation-agnostic electrical connection between the ground station and the charging tether.
Moreover, we supply the ground station with an electromagnet that largely increases the tolerance to localization and control errors during the docking maneuver, while still guaranteeing smooth un-docking once the charging process is completed.
We demonstrate AutoCharge on a perpetual $10$ hours quadrotor flight experiment and show that the docking and un-docking performance is solidly repeatable, enabling perpetual quadrotor flight missions.
\end{abstract}

\section*{Supplementary Material}
\noindent \textbf{Video}: \url{https://youtu.be/6xYvI-qIe3M}

\IEEEpeerreviewmaketitle

\section{Introduction} \label{sec:introduction}
In recent years, unmanned aerial vehicles like quadrotors have drawn significant attention for several applications including search and rescue, transportation, and inspection due to their simplicity in design, agility, low cost, and ability to hover in place and move in 3D~\cite{emran2018reviewquadrotor}.
Nevertheless, these robots are constrained by limited battery endurance which restrains their applicability in persistent, long-distance missions.
The ideal solution for the autonomous charging problem for quadrotors requires a system that is \textit{efficient}, to reduce power waste and heat generation;
\textit{portable}, so that it may be transported and used in different tasks; \textit{universal}, able to charge quadrotors of different frame shapes, sizes, and battery capacities; and \textit{robust}, such that it guarantees persistent docking performance by accommodating large control and localization errors of the quadrotor.

Various solutions have been proposed for extending the flight time of quadrotors, from battery expansion and battery replacement methods~\cite{desilva2022battswap, williams2018battswap, liu2017battswap, ure2015battswap, lee2015battswap, michini2011batterychange}, wireless charging~\cite{jawad2022misalignment, junaid2016wc, choi2016automaticwd, chae2015wc}, contact charging~\cite{jain2020flyingbatteries, cocchioni2014autonomousnl, 2014_auto_charging_vijay, song2014contactcharging}, and tethered charging~\cite{gu2016nuclear, kiribayashi2015tetheredcharging, wang2021hookcharging}.
However, these approaches do not meet all the requirements of the ideal autonomous charging system for quadrotors, but trade-off efficiency, portability, universality, and robustness.

\begin{figure}
    \vspace{1em}
    \centering
    \includegraphics[width=1\linewidth, trim=0 750 0 650, clip]{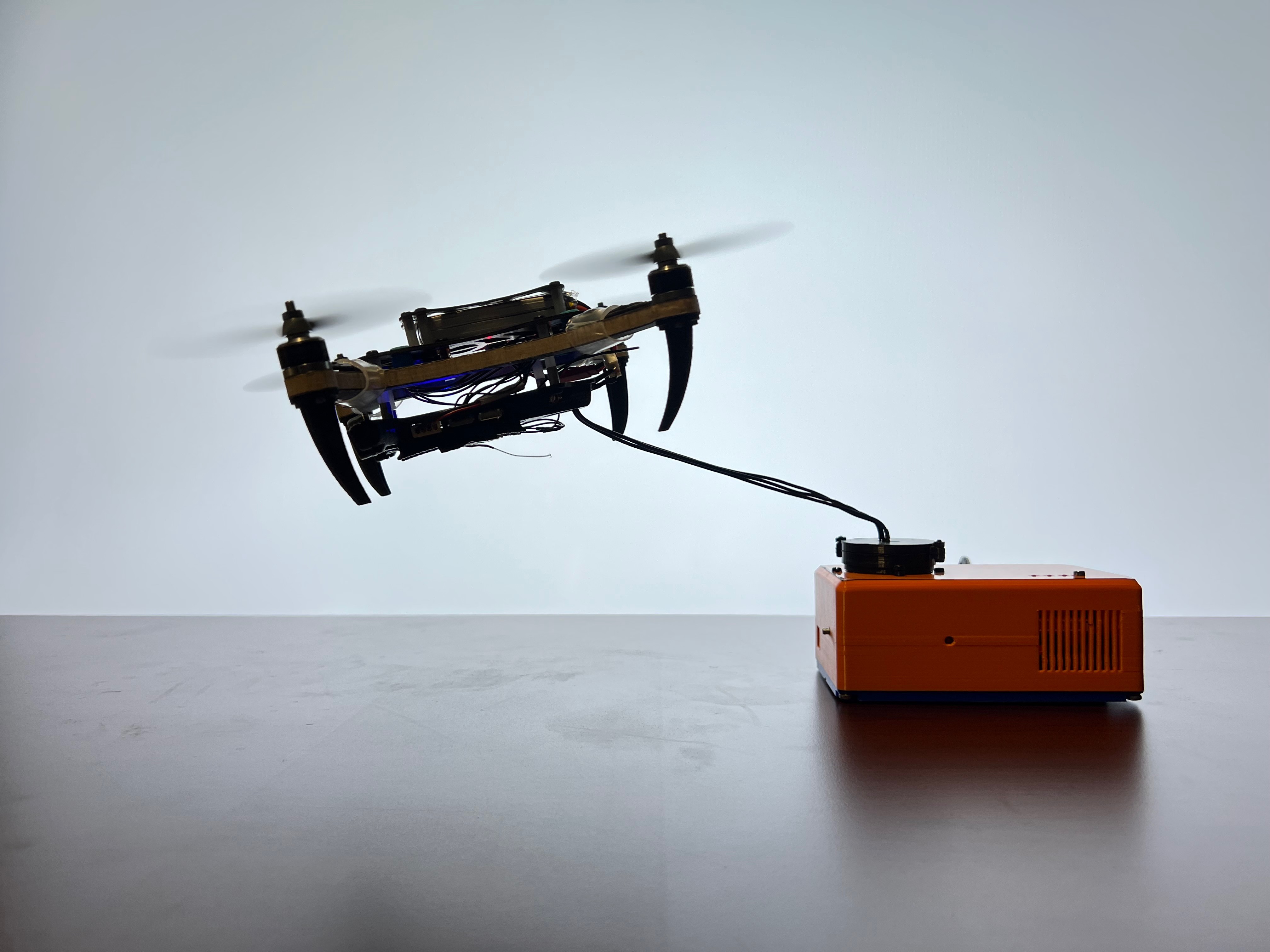}
    \caption{Un-docking maneuver. AutoCharge ensures universal, highly efficient, and robust charging by combining a portable ground station with a flexible, lightweight charging tether.}
    \label{fig:autocharge}
    \vspace{-1.5em}
\end{figure}

In this paper, we propose AutoCharge, an autonomous charging solution for quadrotors that is designed to meet the requirements of the ideal autonomous charging system.
AutoCharge consists of a compact ground station and a flexible charging tether, as shown in Figure~\ref{fig:autocharge}.
The charging is performed through a pair of circular magnetic connectors that establish a precise, orientation-agnostic connection between the tether and the station.
Therefore, by leveraging direct contact charging, AutoCharge ensures low impedance and thus high electrical efficiency while charging.

The ground station is supplied with a powerful electromagnet (EM) to strengthen the magnetic field generated by the connectors. 
The EM is only active during docking and disabled during charging and un-docking.
This guarantees a natural mechanical guide to ensure contact when approaching the ground station, but also an easy and smooth detachment when the charging operation is completed.
Consequently, by leveraging the circular magnetic connectors and the EM, AutoCharge is robust to control and localization errors.
The charging tether acts solely as an additional add-on to the onboard battery, hence introducing minimum quadrotor modifications and enabling AutoCharge to charge quadrotors of different frame shapes and sizes. 
Moreover, the ground station is supplied with a parallel balance charger, enabling the proposed system to target any lithium polymer (LiPo) battery size.
All these characteristics make AutoCharge a universal charging solution.
AutoCharge does not require any reserved area for the quadrotor's body to dock on, as illustrated in Figure~\ref{fig:charging_pipeline}.
As a consequence, the ground station's dimensions are agnostic to the quadrotor's size and the station can be much smaller than the drone making AutoCharge highly portable.

\textbf{Contributions.}
(i) We design and present AutoCharge, an autonomous charging system for quadrotors that consists of a portable ground station and a lightweight, flexible charging tether and is capable of universal, highly efficient, and robust charging;
(ii) We provide a simple and precise description of the manufacturing process used to develop the proposed ground station and charging tether.
Some components of AutoCharge are simple to manufacture from a low-cost ($\sim \hspace{-0.3em} \$300$) 3D printer or milling machine, while others can be directly purchased off the shelf. While commercial solutions available are remarkably expensive, reaching prices up to $\$30$K, AutoCharge's full price does not exceed $\$50$.
(iii) We perform an extensive evaluation of multiple magnet choices to relate their strength and weight to AutoCharge robustness to control and localization errors. Moreover, we validate AutoCharge on a continual $\SI{10}{\hour}$ flight test and show that docking and un-docking operations are smooth and repeatable, enabling perpetual flight missions.

\begin{figure*}[t]
    \centering
    \includegraphics[width=0.95\linewidth, trim=0 350 40 0, clip]{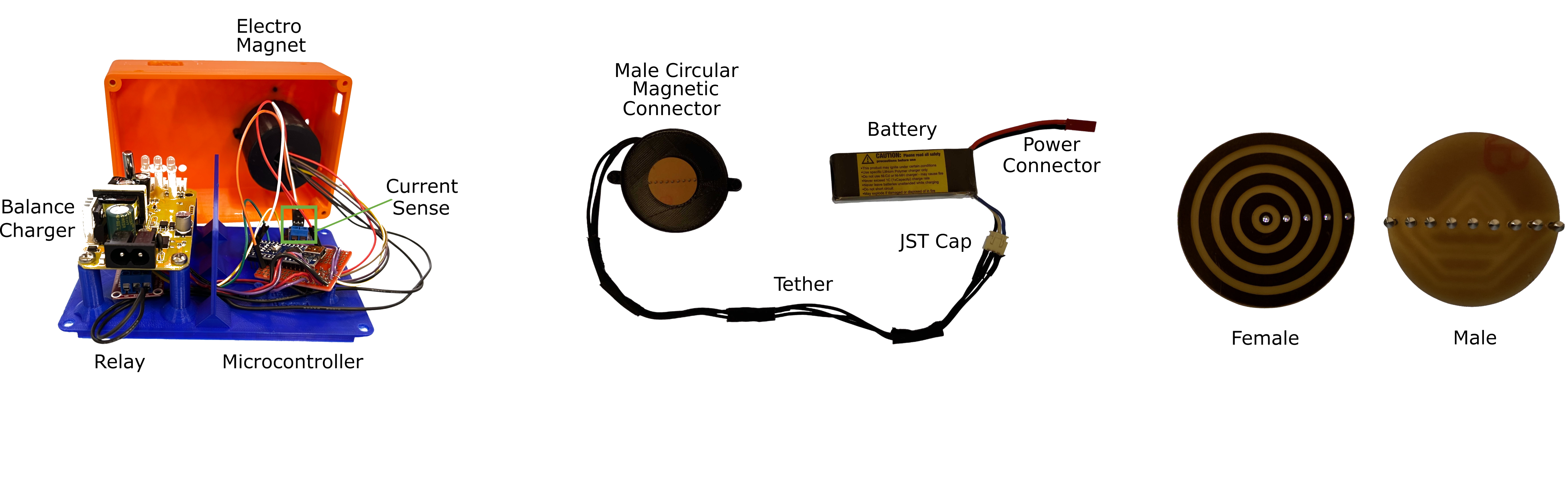}
\end{figure*}
\setlength{\tabcolsep}{5em}
\begin{figure*}[t]
    \vspace{-3em}
    \begin{center}
        \begin{tabular}{c c c}
            (a) Ground Station & (b) Charging Tether & (c) Connectors\\
        \end{tabular}
    \end{center}
    \vspace{-0.5em}
    \caption{AutoCharge consists of a compact ground station (a) and a flexible charging tether (b). The charging is performed through a pair of circular magnetic connectors (c) that establish a precise connection between the tether and the station.}
    \label{fig:manu_components}
    \vspace{-1.5em}
\end{figure*}

\section{Related Works} \label{sec:related_works}

\textit{Battery expansion} represents the simplest option available to increase the quadrotor's mission time.
However, increasing the battery size does not linearly increment the flight time, as demonstrated in~\cite{Leonard_range_estimate}. One of the core reasons is that expanding the battery capacity and size also inevitably increases the weight. Consequently, the motors need to provide more power for lifting and controlling the quadrotor, resulting in more energy being consumed. %This relationship between weight and flight time is shown in Figure~\ref{fig:weight_analysis}.
% However, expanding the battery's capacity and size also implies increasing the weight.
% Consequently, the motors need to provide more power for lifting and controlling the quadrotor, resulting in more energy consumed. 
% In some cases, the additional weight from battery expansion even reduces flight time.
% This relationship between weight and flight time is shown in Figure~\ref{fig:weight_analysis} and formulated in \cite{2014_auto_charging_vijay}.

\textit{Battery replacement} represents a highly efficient solution because it provides the shortest recovery time for a quadrotor to return to flight and can be fully automated through external robotic systems.
However, battery replacement solutions generally include highly-engineered bulky systems that are specifically designed for particular robot structures~\cite{desilva2022battswap, williams2018battswap, ure2015battswap}.
% Moreover, these systems require periodical calibration or feedback control to ensure they remain in phase, hence they are especially prone to jamming.
For example, \cite{michini2011batterychange} proposes a dual-drum structure that holds several batteries and automatically swaps the onboard battery with a charged one. 
Despite being an efficient solution for extending the flight time of quadrotors, the entire system structure is bulky and composed of a tremendous number of components, from microcontrollers and control motors to locking arms and rotational encoders. 
Therefore, the system is not portable and all these components introduce several failure points that may damage the quadrotor and critically interrupt the performed mission.
\cite{lee2015battswap} attempts to simplify the system structure by using fundamental design principles but still does not resolve the problem of failure points during the battery replacement operation.
% Contrarily, AutoCharge is designed to be compact, lightweight, and autonomously charge quadrotors that differ in size and battery capacity.
Another major issue for battery replacement strategies is the need to precisely land on the docking station~\cite{liu2017battswap}. While additional mechanical components can be designed to minimize this problem, this would introduce more complexity and failure points.
% In contrast, AutoCharge fully leverages electromagnetic induction to enable precise docking and un-docking, resulting in minimal moving components and thus guaranteeing high robustness to mechanical failures.
% Finally, battery replacement methods require power cycling the robot. 
% As a result, the software stack is reset and this may cause localization and control degradation once the robot is powered back on.
% Moreover, during charging the quadrotor can continuously sense the environment which may be key for the mission success~\cite{mao2022perching}.
% Conversely, AutoCharge efficiently charges the onboard battery without requiring any power cycle, hence allowing the quadrotor to continuously sense the environment which may be key for the mission success~\cite{mao2022perching}.
In conclusion, battery replacement solutions are \textit{not universal}, \textit{not robust}, and \textit{not portable}.

\textit{Wireless charging} provides a straightforward charging operation that typically only requires introducing a receiver coil on the quadrotor's frame and developing a wireless charging station supplied with a transmitter coil. 
When the coils are close to each other, the onboard battery begins to charge.
For example, \cite{junaid2016wc} presents a charging station using wireless inductive charging, the same technology used for charging smartphones and other electronic devices. 
However, the power transfer efficiency is only about $\SI{75}{\percent}$ when the receiver and transmitter coils are precisely aligned and it significantly degrades even more as the misalignment increases. 
% Achieving precise alignment also requires a very precise landing or a very large receiver coil on the vehicle. 
Several works have sought to address the issues of alignment and poor power transfer efficiency~\cite{chae2015wc}.
% \cite{chae2015wc} proposed a LED system on the charging station to improve the state estimation of the quadrotor. However, this solution requires a relatively long time and additional computation for improving the state information.
For example, the authors in~\cite{choi2016automaticwd} design a wireless charging station that uses ultrasonic sensors for identifying the quadrotor's position after landing. Then, some stepper motors slide the transmitter coil under the quadrotor. As with battery replacement systems, this solution employs multiple mechanical components to coordinate and precisely move, resulting in additional failure points.
Despite the advances in state estimation and mechanical systems for the alignment of the coils, even when these are accurately aligned, wireless charging efficiency remains $\SI{25}{}-\SI{30}{\percent}$ inferior compared to tethered charging solutions~\cite{jawad2022misalignment}, including AutoCharge.
% Therefore, in contrast to battery replacement approaches, wireless charging solutions include compact and portable systems, well generalize to different robot structures, relax the precise landing constraint, and do not require power cycling the quadrotor.
As a result, wireless charging solutions are \textit{not robust} and \textit{not efficient}.

\textit{Contact charging} provides high-efficiency charging by modifying a quadrotor's component, such as the landing gear, to accommodate connectors that establish electrical contact between the vehicle and the charging station after docking~\cite{2014_auto_charging_vijay}.
For example, \cite{song2014contactcharging} proposes new landing gears to host the wires to charge the system as well as a charging station that consists of four metal plates. After landing, the quadrotor is switched off by a weight sensor on the station and the battery gets charged.
% \cite{2014_auto_charging_vijay} designed new landing gears and demonstrated a consistent 9.5 hours of operations through modified landing gears. 
\cite{cocchioni2014autonomousnl} presents similar landing gears with electrical connections from the battery to their lower ends and a passive centering system made of four upside-down hollow cones for correcting the landing positional error.
\cite{jain2020flyingbatteries} shows a mid-air docking and in-flight battery charging approach. A small quadrotor carrying a fully charged battery docks on a bigger quadrotor in mid-air and charges the battery of the latter by using electrical connectors threaded in its landing gear.
Despite the appealing results, contact charging solutions require developing specific quadrotor components for connecting the battery to the external power source, hence not generalizing to different robot structures.
Moreover, these solutions require the quadrotor to precisely land to align the electrical connectors, thus facing control challenges, such as stochastic ground effects or disturbances~\cite{saviolo2023learning, saviolo2022activelearningdynamics}, during docking.
% In contrast, AutoCharge connects the quadrotor to the charging ground station through a lightweight tether, thus only introducing light modifications to the quadrotor's platform and performing the docking maneuver in nominal conditions.
Consequently, contact charging are \textit{not universal} and \textit{not portable}.

\textit{Tethered charging} enables unlimited flight time by directly connecting the quadrotor to a charging station. Hence, this strategy does not need precise physical landing and positioning on a charging station and avoids repeated recharging. 
 \cite{gu2016nuclear} employs tethered charging to perform with a quadrotor a mission in a nuclear power plant.
The major drawback of tethered charging is the flight area that the quadrotor can cover. The charging tether used can not be too long due to the internal resistance and weight of the cable itself which would reduce power efficiency and maneuverability respectively.
Several works have been proposed to overcome this limitation by enabling the ground station to move with the quadrotor.
For example, \cite{kiribayashi2015tetheredcharging} uses an unmanned ground vehicle to carry the ground station that is directly connected to the quadrotor. The vehicle follows the quadrotor and extends the flight area. 
However, by combining aerial and ground vehicles, the quadrotor becomes limited by the ground conditions.
As a result, tethered charging solutions are \textit{not portable}.
The authors in~\cite{wang2021hookcharging} propose a charging system that uses onboard sensing to attach a tether with a pair of loose hooks mid-flight. 
However, this method is not orientation-agnostic because the pair of hooks must match the station's polarity, requires precise control to localize and grasp the tether, and the loose tether attachment limits the quadrotor's ability to roll and pitch to avoid detachment.

\section{Methodology} \label{sec:methodology}
In this section, we introduce the operating principle, key components, and circuit diagram of AutoCharge and describe the manufacturing process performed to fabricate the entire system. The proposed charging system is easy to assemble even by non-experts.
For the sake of clarity and to simplify the design, we manufacture the components of AutoCharge for charging up to 4S LiPo batteries. However, the same manufacturing process can be extended to LiPo batteries of larger capacities by including more copper rings in the connectors.
Figure~\ref{fig:manu_components} illustrates the manufactured components of AutoCharge. The 3D printed components were designed in SolidWorks and manufactured through a low-cost Chiron 3D printer, while the circuit components were designed in EAGLE and fabricated through an OtherMill Pro.

\subsection{Operating Principle}
AutoCharge's operating principle is illustrated in Figure~\ref{fig:charging_pipeline}.
When the onboard battery is running low, the quadrotor approaches the charging station and the natural magnetic force generated by the EM precisely auto-aligns the connectors.
Once the electrical connection is established, the EM is deactivated, the charging operation begins, and the quadrotor lands.
During charging, the quadrotor's software stack remains active and no power cycling occurs. 
This guarantees that while refueling the quadrotor can perform multiple secondary mission tasks~\cite{mao2022perching, morando2022inspection, morando2020uavpatrol}.
When the charging operation is completed, the quadrotor smoothly un-docks from the ground station and continues the mission.

\subsection{Ground Station}
The ground station (Figure~\ref{fig:manu_components}a) is designed to enable efficient charging once the electrical connection with the charging tether is established. 
The station is mounted to the ground and attached to an external power source. 
The key components of the station are an electrical circuit (Section~\ref{sec:circuit}), an EM, a female circular magnetic connector, and a poly-lactic acid (PLA) enclosure.
The EM generates a powerful magnetic field that attracts the magnetic head of the charging tether when the quadrotor is approaching the station. 
The magnetic force is then switched off during charging and un-docking.
This design ensures a fast, robust docking procedure along with smooth un-docking.

The ground station is designed to be flexible and adapt to different flight operations, hence trading-off between portability and robustness.
For example, if the mission is carried out in an outdoor environment characterized by stochastic wind effects that degrade the control and localization performance, then it is key to strengthen the magnetic field generated by the EM.
Contrarily, if the flight mission is performed indoor with relatively accurate state estimation and control algorithms, then portability can be maximized by employing a smaller-scale EM.

\textbf{Manufacturing.} 
The female circular magnetic connector (Figure~\ref{fig:manu_components}c) is manufactured in-house through a \si{mm} level precision (printed circuit board) PCB mill. A through hole is added to the female connector to allow electric connection from the back. The ground station enclosure and fasteners, which are used for aligning and holding the female connector and other electronic components, are 3D printed and assembled by simply screwing the appropriate parts together visible in Figure~\ref{fig:manu_components}a.
Overall the manufactured ground station weights $\SI{0.56}{\kilo\gram}$ and has the dimensions $15 \times 10 \times 6 ~\si{cm^3}$.

\subsection{Charging Tether}
The charging tether (Figure~\ref{fig:manu_components}b) is a custom cable that remains always connected to the battery, dangling down the quadrotor's frame during flight operation. 
The cable consists of a low-resistance 20 gauge multi-core wire that connects a male JST cap to a male circular magnetic connector.
The tether's dangling head is supplied with a male line of pogo pins magnetic connector that matches the female circular connector on the ground station.
The male connector is designed to be slightly concave, ensuring that while docking the electrical connection is established only when the male and female circuits perfectly mate, thus avoiding potential dangerous shorting issues.

The charging tether's length is arbitrary and should be chosen based on the carried flight task. 
If during charging the quadrotor is passively waiting for the operation to be completed, then the tether's length should be chosen short to minimize the effect on the dynamics of the system and minimize efficiency loss from increased resistance from a longer tether. 
Contrarily, if the quadrotor is required to perform active tasks during charging, such as inspection or surveillance, then the tether's length can be relatively long.
We refer to recent works on tethered flight for a detailed study on the choice of the tether's length and how the cable's resistance affects the charging efficiency~\cite{kiribayashi2015tetheredcharging,vishnevsky2019principles,jain2022tetheredanalysis}.
The charging tether's weight is mainly dominated by the weight of the magnetic connector.
The magnetic strength of this connector can be fully customized for the considered application.
Hence, trading-off portability and robustness is analogous to the EM design choice.
We explore this trade-off in detail in the proposed experiments in  Section~\ref{sec:exp_results}.

\textbf{Manufacturing.}
Our tether is composed of $20~\si{Gauge}$ multi-core wires which connect both our battery connector to the ground station. 
The battery connector is a JST soldered onto one end of the wires and hosts on the other end of the wires the male circular magnetic connector that establishes the electrical connection with the ground station.
The connector is composed of a circular magnet and a set of pogo pins.
We design and 3D print a PLA enclosure that can contain both the electromagnet and pogo pins and secure them in place. 

\begin{figure}[t]
    \centering
    \includegraphics[width=0.9\linewidth]{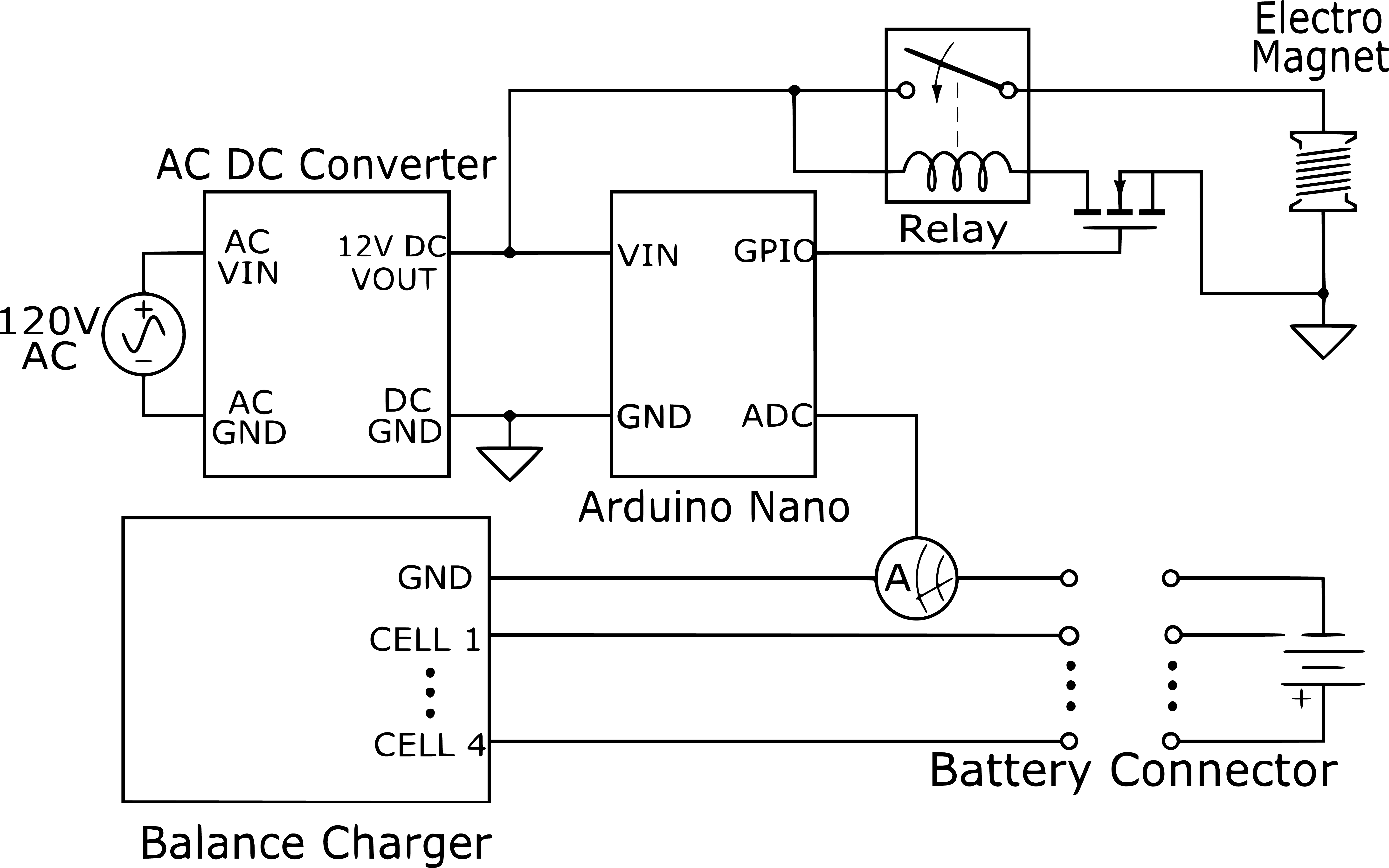}
    \caption{AutoCharge's circuit diagram.}
    \label{fig:circuit_schematic}
    \vspace{-1.5em}
\end{figure}

\subsection{Circuit Diagram} \label{sec:circuit}
AutoCharge's circuit diagram (Figure~\ref{fig:circuit_schematic}) is composed of a balance charger and an EM control circuitry.
The balance charger is directly connected to the female magnetic circular connector which mates with the battery.
This circuit block automatically detects and supplies power to the attached number of cells, hence providing a universal, efficient, and balanced charging operation up to $4$ cells. The number of cells can be scaled up arbitrarily.  
The EM circuit is controlled through an Arduino Nano microcontroller and powered through an AC-DC converter rail, allowing charging operations anywhere nearby a power socket.
A relay controls the switching action of the EM.
% The EM's magnetic field does not affect the relay's switching mode based on experimental validation.
In idle conditions, while the quadrotor is not attached, the relay closes and the current flows allowing the electromagnet to pull the tether. 
The microcontroller detects battery attachment by measuring the amount of current flowing through the battery connector and switches the relay open shutting down the EM. The vehicle can measure its internal battery voltage to estimate its current capacity and take off autonomously once a sufficient amount of charge has been accumulated. 
After the quadrotor takes off, no current flows through the connector and the relay closes after a short delay allowing another charging iteration to occur.
This provides both the benefit of robust docking from high magnetic fields and easy detach operations. An additional wireless communication device can be implemented to remotely control the EM for greater control.

\textbf{Manufacturing.}
Each electronic block (Arduino Nano, relay, balance charger, converter, and EM), aside from the magnetic connector, were purchased off-the-shelf. All the components are electrically connected through soldering. 

\begin{figure*}[t]
    \centering
    \includegraphics[width=0.9\linewidth, trim=0 530 0 0, clip]{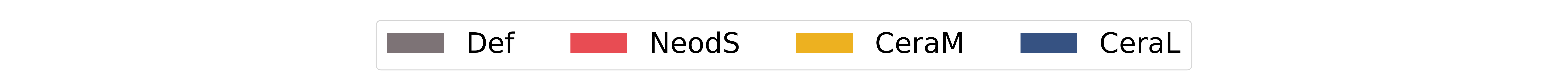}
    \includegraphics[width=0.95\linewidth, trim=0 0 0 0, clip]{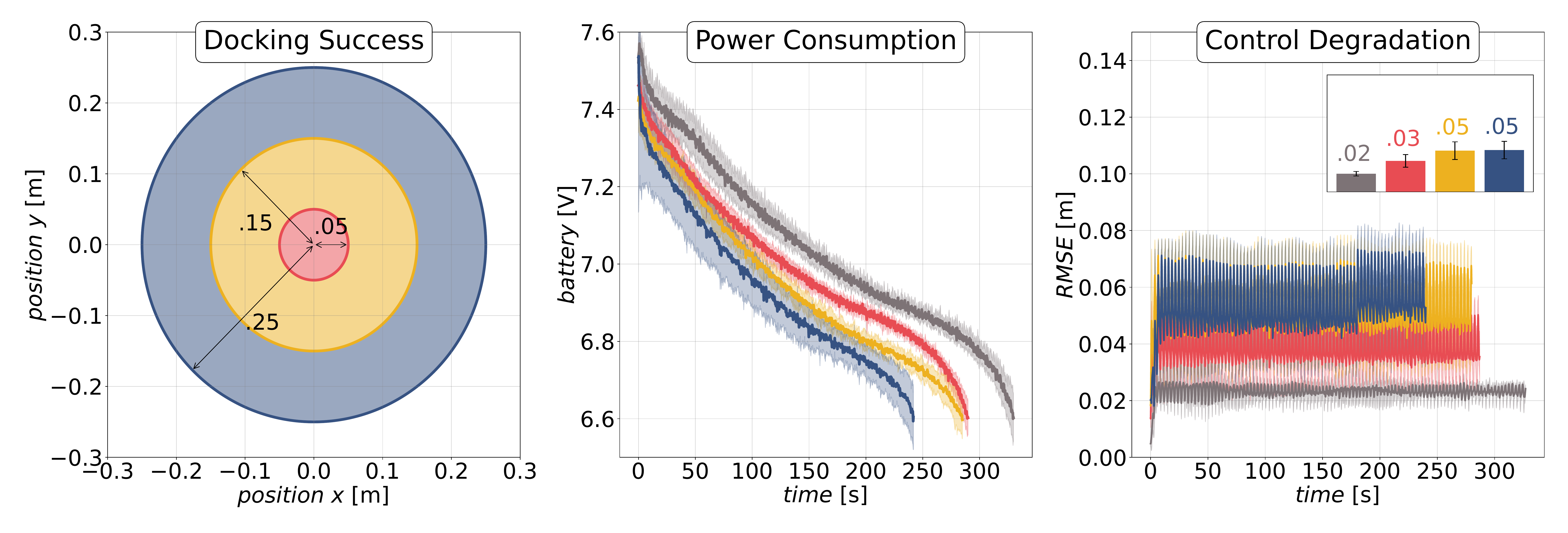}
    \hfill
    \vspace{-1em}
    \caption{Impact of different choices of magnetic connectors. (Inset) Average RMSE.}
    \label{fig:weight_analysis}
    \vspace{-1.5em}
\end{figure*}

\section{Experimental Setup} \label{sec:exp_setup}
We validate the robustness of AutoCharge by running multiple experiments in both indoor and outdoor environments. Specifically, the indoor experiments are conducted at the Agile Robotics and Perception Lab (ARPL) lab at New York University flying arenas, and the outdoor experiments are performed on a rooftop terrace.
The indoor flying arena is equipped with a Vicon motion capture system which provides accurate state estimates at $100~\si{Hz}$.
For outdoor flights, visual-inertial odometry algorithm combined with IMU measurements with an unscented kalman filter provides state estimates at $500~\si{Hz}$ and controlled using a nonlinear controller based on our previous work~\cite{loianno2016estimation}.
Trajectories are planned using trapezoidal velocity profiles.
We compare different design choices of AutoCharge and evaluate the trade-off between portability and robustness introduced in Section~\ref{sec:methodology}.
Specifically, we alter the quadrotor's default configuration (Def) with three charging tethers of length $0.5~\si{m}$ with different male magnetic connectors: a small neodymium magnet of weight $0.42~\si{g}$ and pulling force $771.11~\si{g}$ (NeodS), a medium ceramic magnet of weight $17.5~\si{g}$ and pulling force $2721.55~\si{g}$ (CeraM), and a large ceramic magnet of weight $34.7~\si{g}$ and pulling force $4989.52~\si{g}$ (CeraL).
We demonstrate the universality of AutoCharge by using two quadrotors of different frame sizes, battery capacities, and thrust-to-weight ratios for conducting the experiments.
The first quadrotor is equipped with a Qualcomm\textsuperscript{\textregistered} Snapdragon\textsuperscript{TM} board and four brushless motors and weights $250~\si{g}$ including the battery. 
This quadrotor is charged by a $2$-Cell/$2$S battery with a capacity of $910~\si{mAh}$ that weights $47~\si{g}$ and has a maximum voltage of $7.4$~\si{V}.
The second quadrotor is equipped with a Nvidia\textsuperscript{\textregistered} Jetson Xavier\textsuperscript{TM} NX board and four brushless motors and weights $890~\si{g}$ including the battery. 
This quadrotor is equipped with a $4$-Cell/$4$S battery with a capacity of $3000~\si{mAh}$ that weights $281~\si{g}$ and has a maximum voltage of $14.8$~\si{V}. We use \textit{SD2S} and \textit{NX4S} to refer to the lighter and heavier quadrotor respectively.

\section{Results} \label{sec:exp_results}
We design our evaluation procedure to address the following questions.
(i) What is the impact of the charging tether's weight for different choices of magnet on the docking success, power consumption, and control performance?
(ii) Can AutoCharge be employed to autonomously charge quadrotors with various frame shapes and battery capacities?
(iii) Does the proposed system enable perpetual autonomous charging in a long flight mission?
We encourage the reader to watch the multimedia material for additional qualitative results.

\subsection{Portability vs Robustness}
We investigate the impact of different choices of magnetic connectors on the docking success, power consumption, and control degradation of the SD2S quadrotor.
We evaluate the docking success in terms of the maximum distance from which the ground station pulls the male magnetic connector.
Moreover, we compare the power consumed and the control degradation when using different tether configurations to continuously track a circular trajectory of radius $\SI{1}{\meter}$ at $\SI{2}{\meter/\second}$ until the battery voltage reaches $\SI{6.6}{\volt}$.
The control degradation is evaluated based on the root mean squared error (RMSE) between the quadrotor position and reference trajectory at every control iteration and the power consumption in terms of battery voltage over time.
The experiments are repeated $5$ times to estimate the mean and standard deviation of both metrics.
For each experiment, the quadrotor's mass is scaled appropriately for the controller, and the ground station's EM and magnetic attractiveness remain constant. 

\begin{figure*}[t]
    \centering
    \includegraphics[width=\linewidth, trim=0 85 0 0, clip]{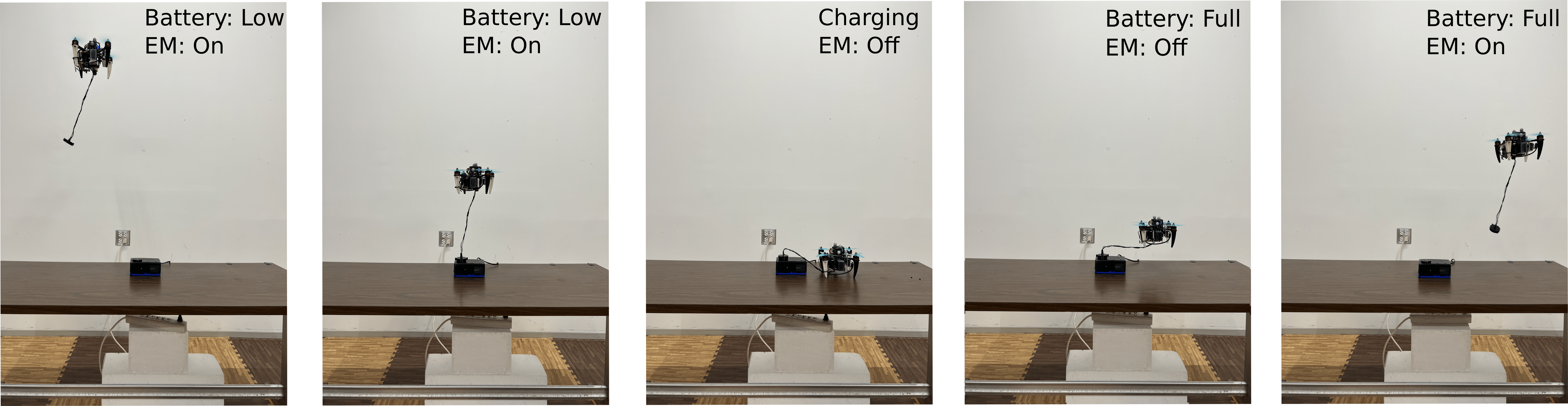}
    \captionof{figure}{
    Docking and un-docking performance in the indoor environment with NX4S quadrotor.}
    \label{fig:universality_analysis}
    \vspace{-1.0em}
\end{figure*}

\begin{figure*}[t]
    \centering
    \includegraphics[width=\linewidth]{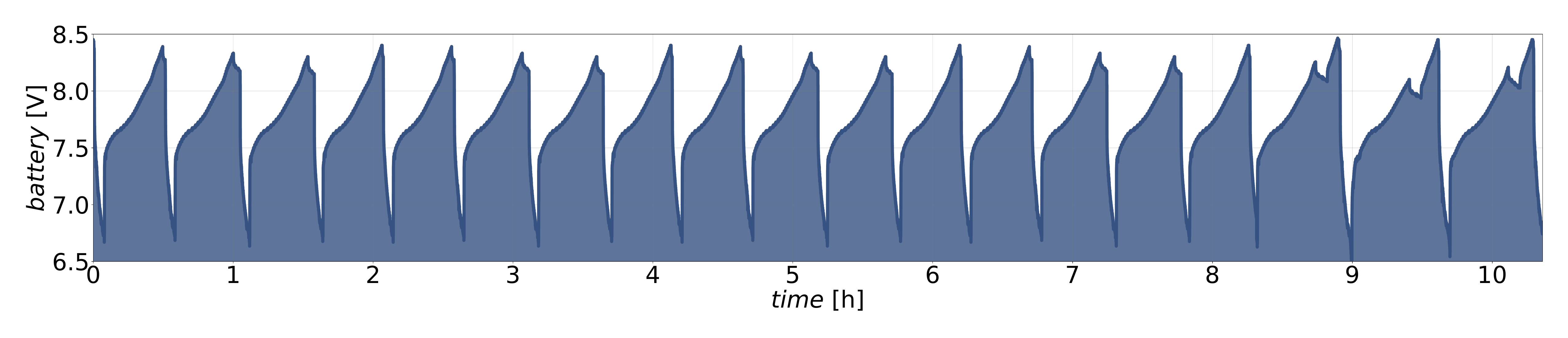}
    \vspace{-2.0em}
    \caption{Battery voltage over time during a fully autonomous persistent flight mission.}
    \label{fig:perpetual_flight}
    \vspace{-1.5em}
\end{figure*}

Figure~\ref{fig:weight_analysis} illustrates the results of this experiment.
The additional weight increases the amount of thrust that the motors need to provide for lifting and controlling the quadrotor. Consequently, the flight time for heavier magnetic connectors is inferior to smaller ones, resulting in a flight time degradation by up to $15 \%$.
Moreover, the results show that altering the quadrotor's system with the lighter charging tether does not significantly affect the tracking performance. 
Therefore, this demonstrates that the proposed charging solution does not significantly alter the quadrotor's system dynamics.
Importantly, the results show that employing larger and stronger magnets directly impacts the docking success, by improving the pull distance by $\times 5$. 
This boosted docking performance may be critical for applications where localization and control errors are unavoidable (e.g., outdoor environments affected by stochastic wind gusts), enabling the quadrotor to reliably perform precise docking operations.
When the male circular magnetic connector is within the bounds of the docking success area, the attachment operation had a $100 \%$ success rate in all the performed experiments.
We further study and validate this performance by controlling a quadrotor to continuously attach and detach from the ground station over $100$ iterations.
AutoCharge enables solidly repetitive attachment and detachment.
We refer to the supplementary video for a qualitative demonstration.

\subsection{Universality Analysis}
Universality is a desirable characteristic of any charging system.
Every system should demonstrate the capability to autonomously charge different quadrotor frame sizes and battery capacities.
Therefore, we study AutoCharge's ability to autonomously charge the quadrotors SD2S and NX4S. %, previously introduced.
Specifically, we control the quadrotors to repetitively perform the docking and un-docking operations to simulate the charging process during perpetual missions. Figure~\ref{fig:charging_pipeline} and Figure~\ref{fig:universality_analysis} illustrate some snapshots of this experiment.
We encourage the reader to refer to the supplementary multimedia material for additional performances of both quadrotors.
The results show that the docking and un-docking performance is solidly repeatable while using the same connectors for different quadrotors with 2S and 4S batteries, hence validating AutoCharge as a universal charging solution.

\subsection{Perpetual Quadrotor Flight}
We demonstrate the ability and flight time benefits of employing AutoCharge on a long perpetual flight test.
Specifically, we employ the quadrotor SD2S to track multiple trajectories until the battery voltage reaches $\SI{6.6}{\volt}$. Then, the quadrotor is required to reach the ground station, dock, and recharge.
After charging is complete, the quadrotor detaches from the ground station and continues tracking the random trajectories.
The experiment ends after $\SI{10}{\hour}$.

Figure~\ref{fig:perpetual_flight} illustrates how the battery voltage changes over time during the entire experiment. 
The quadrotor consistently and robustly docks, charges, and un-docks for long periods without any human intervention.
Moreover, the results show no noticeable battery degradation over the entire flight, hence validating AutoCharge for safe and efficient charging for quadrotors. 
Towards the end of the $10\si{h}$ flight, the charger's temperature protection is triggered causing it to throttle the charging current till the ideal operating temperature range is reached. Successively, the charging operation is resumed and the battery is charged until completion. This behavior creates short voltage dips that characterize the last voltage peaks.

\section{Discussion and Limitations} \label{sec:conclusion}
Autonomously charging has the potential to staggeringly empower future applications for quadrotors, such as expanding the range of delivery systems, persistently inspecting large crop fields to identify pests, and acting as a mobile communication hub during disaster management.
Commercial solutions available do not satisfy the requirements of the ideal autonomous charging solution and are terribly expensive, reaching prices up to $\$30$K~\cite{dji2022,hextronics2022,idiployer2022,skycharge2022}.
In this paper, we proposed AutoCharge, an autonomous charging system for quadrotors that is capable of universal, highly efficient, and robust charging.
We validated these capabilities in several experiments where AutoCharge demonstrated high flexibility to different quadrotors, battery capacities, system dynamics, and task objectives.
Moreover, we stress-tested AutoCharge for over $10$ hours to validate its charging repeatability.

AutoCharge offers a highly-flexible charging solution that can be customized to the considered application.
Specifically, larger stations can employ stronger magnets allowing less accurate control to dock with the station. This magnet force increase comes at the cost of less portable stations and more external forces on the vehicle.
Future work will tackle this problem by modeling the charging tether as a cable suspended payload~\cite{guanrui2021pcmpc} and developing an admittance controller to accommodate large magnetic forces~\cite{ott2009admittance} creating a smooth transition for the quadrotor during the docking maneuver.
% Because the magnetic force is known as a function of the magnet size and position with few confounding factors to reduce the force in the wild, admittance control can be possibly achieved without any additional force sensors. %Future works can explore the trade-offs between weight and stability between adding such a sensor or mathematically calculating the virtual force. 

Future works will also focus on boosting the usability of the proposed charging solution without prior knowledge of the location of the ground station, but using cameras to visually localize it and control the quadrotor in an image-based visual servoing fashion~\cite{thomas2014ibvs}.

\clearpage

\bibliographystyle{IEEEtran}
\bibliography{references}

\end{document}